%% file: acl2023.tex
\pdfoutput=1

\documentclass[11pt]{article}

\usepackage{ACL2023}

\usepackage{times}
\usepackage{latexsym}
\usepackage{tikzsymbols}

\usepackage[T1]{fontenc}

\usepackage[utf8]{inputenc}

\usepackage{microtype}

\usepackage{inconsolata}
\usepackage{bbm}
\usepackage{todonotes}
\usepackage{multirow}
\usepackage{arydshln}
\usepackage{booktabs}
\usepackage{amsmath}

\usepackage{listings}
\usepackage{xcolor}
\newcommand{\sysname}{\textsc{API-Blend}}

\colorlet{punct}{red!60!black}
\definecolor{background}{HTML}{EEEEEE}
\definecolor{delim}{RGB}{20,105,176}
\colorlet{numb}{magenta!60!black}

\lstdefinelanguage{json}{
    basicstyle=\scriptsize\ttfamily,
    numbers=none,
    numberstyle=\scriptsize,
    stepnumber=1,
    numbersep=8pt,
    showstringspaces=false,
    breaklines=true,
    frame=lines,
    backgroundcolor=\color{background},
    literate=
     *{0}{{{\color{numb}0}}}{1}
      {1}{{{\color{numb}1}}}{1}
      {2}{{{\color{numb}2}}}{1}
      {3}{{{\color{numb}3}}}{1}
      {4}{{{\color{numb}4}}}{1}
      {5}{{{\color{numb}5}}}{1}
      {6}{{{\color{numb}6}}}{1}
      {7}{{{\color{numb}7}}}{1}
      {8}{{{\color{numb}8}}}{1}
      {9}{{{\color{numb}9}}}{1}
      {:}{{{\color{punct}{:}}}}{1}
      {,}{{{\color{punct}{,}}}}{1}
      {\{}{{{\color{delim}{\{}}}}{1}
      {\}}{{{\color{delim}{\}}}}}{1}
      {[}{{{\color{delim}{[}}}}{1}
      {]}{{{\color{delim}{]}}}}{1},
}

\title{\sysname{}: A Comprehensive Corpora for Training and \\ Benchmarking API LLMs}

\author{ 
        Kinjal Basu$^*$,
        Ibrahim Abdelaziz$^*$,
        Subhajit Chaudhury, 
        Soham Dan, \\
        {\bf Maxwell Crouse,
        Asim Munawar,
        Sadhana Kumaravel,
        Vinod Muthusamy,} \\
        {\bf Pavan Kapanipathi, and Luis A. Lastras}\\ 
         IBM Research  \\
        {\normalsize{ \{kinjal.basu, ibrahim.abdelaziz1, subhajit, soham.dan, maxwell.crouse, asim\}@ibm.com}} \\
        {\normalsize{ sadhana.kumaravel1@ibm.com, \{vmuthus, kapanipa, lastrasl\}@us.ibm.com}}
    }

\begin{document}
\maketitle
\def\thefootnote{*}\footnotetext{These authors contributed equally to this work}\def\thefootnote{\arabic{footnote}}

\begin{abstract}


There is a growing need for Large Language Models (LLMs) to effectively use tools and external Application Programming Interfaces (APIs) to plan and complete tasks. As such, there is tremendous interest in methods that can acquire sufficient quantities of train and test data that involve calls to tools / APIs. Two lines of research have emerged as the predominant strategies for addressing this challenge. The first has focused on synthetic data generation techniques, while the second has involved curating task-adjacent datasets which can be transformed into API / Tool-based tasks. In this paper, we focus on the task of identifying, curating, and transforming existing datasets and, in turn, introduce \sysname{}, a large corpora for training and systematic testing of tool-augmented LLMs.
The datasets mimic real-world scenarios involving API-tasks such as API / tool detection, slot filling, and sequencing of the detected APIs. We demonstrate the utility of the \sysname{} dataset for both training and benchmarking purposes\footnote{\sysname{} data generation code can be accessed here: \url{https://github.com/IBM/API-BLEND}}. 

\end{abstract}

\input{sections/intro_subho}
\input{sections/2-related-work}

\input{sections/3-dataset-curation-soham}

\input{sections/5-experiments}
\section{Conclusion}
This paper presents \sysname{}, a large corpora for training and evaluation of tool-augmented LLMs, curated from real-world datasets of different domains such as dialog, semantic parsing, digital assistants, etc. \sysname{} consists of 10 datasets (5 in-distribution and 5 out-of-distribution) comprising over 190k instances (including train, development, and test). We have demonstrated that the models fine-tuned with \sysname{} generalize better than the other tool-augmented LLMs on OOD experiments. Our findings not only substantiate the importance of \sysname{} in training and benchmarking tool-augmented LLMs but also highlight the necessity of generalizability to improve the API usage capability of LLMs. 


\input{sections/limitations}

\bibliographystyle{acl_natbib}
\bibliography{references}  

\end{document}

%% file: sections/intro_subho.tex
\section{Introduction}

Large Language Models (LLMs) have shown remarkable abilities across a variety of Natural Language Understanding (NLU) tasks \cite{min2023recent}, e.g., text generation~\cite{brown2020language, radford2019language}, summarization~\cite{zhang2020pegasus, beltagy2020longformer}, and mathematical reasoning~\cite{imani2023mathprompter}. 
There has been strong recent interest in enabling LLMs to call APIs or external tools (such as calculators, calendars, or web searches \cite{hao2023toolkengpt,qin2023toolllm,toolalpaca}) to accomplish high level tasks like booking a hotel, reserving a table, and automating a job requisition tasks. These higher-level tasks are generally conversational and complex. However, in order to perform such complex tasks, LLMs should be able to perform simpler tasks with APIs such as (a) APIs detection: Based on a user query, correctly choose which API to call, (b) Slot filling\footnote{In this paper, \textit{slot} and \textit{input parameters} are used interchangeably.}: Given the API, extract either the slots/parameters from the user utterances or request from the user more information to fill the required parameters of the detected API, and (c) Sequencing: Given an utterance specifying a task, write the sequence of APIs that needs to be called to accomplish the task.


 \begin{figure}[t]
 \begin{center}
 \includegraphics[width=0.5\textwidth]{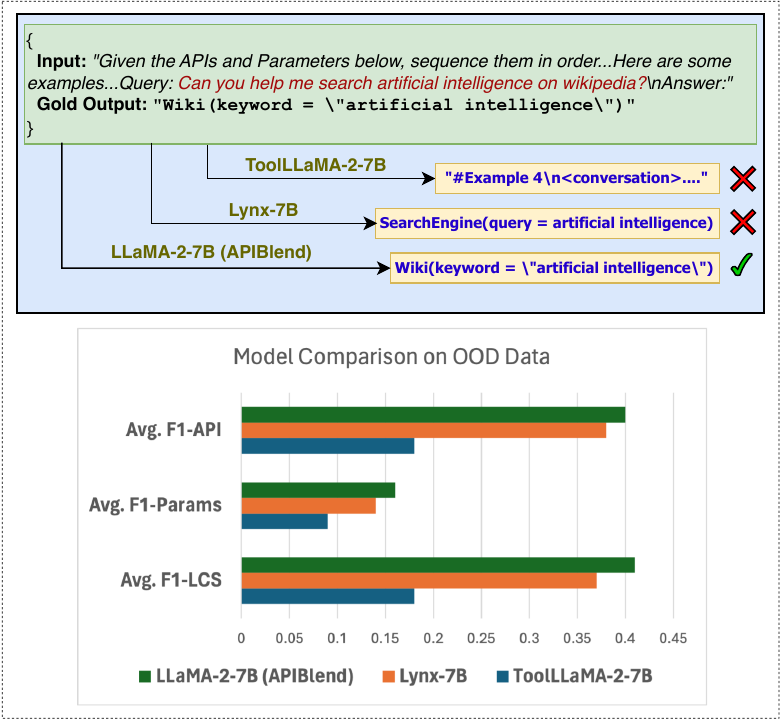}
 \caption{\textit{Top}: an example from the API Bank-Level1 dataset (OOD) that showcases LLaMA-2-7b fine-tuned with \sysname{} generates the correct API and parameter, whereas the other models hallucinate. \textit{Bottom}: performance comparison among three models of similar sizes; two recent tool-augmented models (Lynx and ToolLLaMA-2-7B) and a LLaMA-2-7B model trained with \sysname{} (\sysname{}-LLaMA-2-7B), which significantly outperforms the other two models.}
 \label{fig:toolpile_example}
 \end{center}
 \end{figure}

Data for the above-mentioned API tasks, both for training and benchmarking LLMs has been scarce. Addressing this issue, in the recent past, most approaches have relied on synthetically generated data for training API-augmented LLMs. For instance, ToolLLM \cite{qin2023toolllm} produces multi-sequence REST-API data sourced from GPT-4 \cite{achiam2023gpt}, while datasets like Gorilla \cite{patil2023gorilla} utilize synthetic, single-sequence API data, specifically Deep Learning libraries' APIs, generated from language models. 

Although generating synthetic data from LLMs offers a cost-effective means of obtaining substantial training data, they suffer from several drawbacks. First, data generation methods are plagued by critical issues such as bias inherent in the sampling model, and a lack of diversity in the training dataset. Previous work has shown synthetically generated data suffers from lack of diversity \cite{gupta2023targen}, and diverse data can improve out-of-domain (OOD) generalization \cite{yu2022can}. 
Consequently, models trained on such data can overfit on in-distribution APIs and struggle to generalize to OOD APIs that were not encountered during training, restricting the broad applicability of LLMs for tool usage. 
In addition, datasets have primarily included API detection (single and multiple) and Slot filling in different settings, whereas Sequencing, a prominent task to perform higher-level human tasks using APIs has rarely been the focus in existing works. Lastly, datasets in domains such as digital assistants and semantic parsing that are related to API-tasks and are human-annotated have gone unnoticed in literature due to the emergence of synthetic data generation techniques; 

\begin{figure}[tb]
\begin{center}
\includegraphics[width=0.5\textwidth]{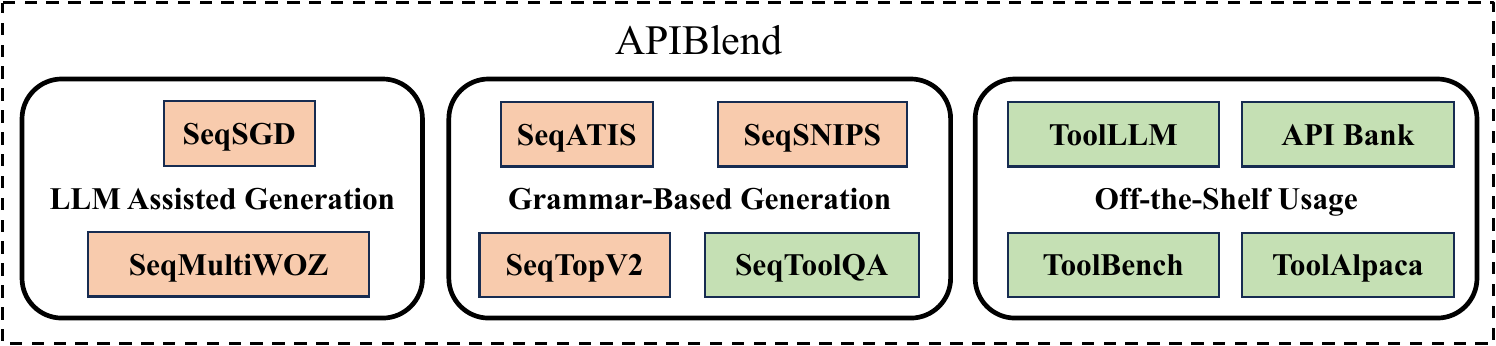}
\caption{\sysname{} Datasets: 10 datasets, 6 curated as part of this paper and 4 are off-the-shelf datasets used for out-of-domain testing. }
\label{toolpile}
\end{center}
\end{figure}

In light of the above issues, we have developed an API-focused training dataset that leverages a hybrid approach for data generation. This is built upon five human-annotated datasets with LLM-assisted generation comprising of over 150k, 17k, and 11k train, development, and test instances. The transformation primarily focuses on sequencing, including API detection and slot-filling due to the sparsity and importance of sequencing data for training models. Furthermore, these datasets are collected from different domains such as semantic parsing, dialog, and digital assistants resulting in a higher diversity of API data. We show that models trained on this diverse dataset yield significantly better OOD generalization performance compared to other state-of-the-art methods, with an example shown in Figure \ref{fig:toolpile_example}. As an evaluation/benchmarking dataset, we include five different existing benchmarks for OOD evaluation. In conclusion, we release API-Blend, a comprehensive API training, and benchmarking dataset, comprising of 10 datasets (5 for training, and 5 for OOD testing), see Figure \ref{toolpile}.

%% file: sections/2-related-work.tex
\section{Related Work}

\subsection{Tool-Usage by LLMs}

Many recent works \cite{komeili2022internet,thoppilan2022lamda,gao2023pal,Schick2023ToolformerLM} have explored how to address the susceptibility of current LLMs to certain errors (e.g., arithmetic \cite{patel-etal-2021-nlp}) through the use of external tools. Such tools can be called by an LLM to provide itself with access to up-to-date information \cite{komeili2022internet,Schick2023ToolformerLM}, perform mathematical operations \cite{he2023solving}, and even execute formal programs \cite{gao2023pal}.

Early approaches to general-purpose training of LLM tool-use leveraged large amounts of human-annotated data \cite{komeili2022internet,thoppilan2022lamda}. The difficulty in scaling these approaches was addressed by later works, which utilized self-supervision \cite{Schick2023ToolformerLM,parisi2022talm} and few-shot prompting \cite{yao2022react}. The prompting framework of \cite{yao2022react} has become widely used when augmenting LLMs with tools, with many follow-up works exploring how to improve its cost-effectiveness \cite{xu2023rewoo}, performance \cite{shinn2023reflexion,yang2023mm}, and data generation quality \cite{qin2023toolllm,tang2023toolalpaca}.

The utility of tool-calling itself has been explored with many standard benchmarks for question answering \cite{saikh2022scienceqa,yang2018hotpotqa}, mathematical reasoning \cite{cobbe2021training}, machine translation \cite{scarton2019estimating,lewis2020mlqa}, and planning \cite{shridhar2020alfworld}. While useful to serve as a comparison against task-specific, supervised methods, it is unclear to what extent these datasets actually require the usage of tools. As observed by \cite{zhuang2023toolqa}, such benchmarks do not adequately distinguish between problems that can be solved using only an LLM's internal knowledge and those that can only be solved through tool calls.


\subsection{API Datasets}

The first self-supervised approaches to constructing tool-use datasets \cite{Schick2023ToolformerLM,parisi2022talm} focused on a small set of general-purpose tools. Soon after, tool-use was quickly expanded to general API function calling \cite{qin2023toolllm,tang2023toolalpaca,patil2023gorilla}, where the volume and diversity of APIs and scenarios were instead emphasized. While all of the aforementioned datasets highlight the number of APIs involved in their respective corpora, they each vary in terms of how those API calls are utilized. For instance, some datasets curate scenarios involving only a single API call \cite{tang2023toolalpaca,patil2023gorilla,xu2023tool} while others involve multiple calls \cite{qin2023toolllm,hao2023toolkengpt}. In addition, some require actual calls to a real API to solve their problems \cite{qin2023toolllm,li2023api,xu2023tool}, which contrasts with other works that simulate API calls with a prompted LLM \cite{tang2023toolalpaca,patil2023gorilla}.

A limitation of the above-listed self-supervised corpora lies in the evaluation of API-use scenarios. Some approaches evaluate based on hallucination rate \cite{patil2023gorilla} while others rely on a separate LLM to assess the quality of an example \cite{tang2023toolalpaca,qin2023toolllm}. Recent works have focused on this issue, with \citet{farn2023tooltalk} relying on smaller sets of manually collected ground truth annotations and \citet{huang2023metatool} performing manual inspection of generated data.

%% file: sections/3-dataset-curation-soham.tex
\begin{table}[]
  \centering
  \scriptsize
\begin{tabular}{l|ccr|rr}
\toprule
\textbf{Datasets}                            & \multicolumn{1}{l}{\textbf{Train}}  & \multicolumn{1}{l}{\textbf{Dev}}   & \multicolumn{1}{l|}{\textbf{Test}} & \multicolumn{1}{l}{\textbf{\begin{tabular}[c]{@{}l@{}}Avg.\\  Seq.\\ Len\end{tabular}}} & \multicolumn{1}{l}{\textbf{\begin{tabular}[c]{@{}l@{}}Avg. No. \\ Params\end{tabular}}} \\ \toprule
{\color[HTML]{CB0000} \textbf{SeqATIS}}      & \multicolumn{1}{r}{11,670}           & \multicolumn{1}{r}{694}            & 774                                & 2.13                                                                                    & 4.85                                                                                     \\
{\color[HTML]{CB0000} \textbf{SeqSGD}}       & \multicolumn{1}{r}{6,782}            & \multicolumn{1}{r}{1,092}           & 1,567                               & 2.44                                                                                    & 3.5                                                                                      \\
{\color[HTML]{CB0000} \textbf{SeqSNIPS}}     & \multicolumn{1}{r}{39,750}           & \multicolumn{1}{r}{2,198}           & 2,199                               & 1.96                                                                                    & 5.06                                                                                     \\
{\color[HTML]{CB0000} \textbf{SeqMultiWOZ}}  & \multicolumn{1}{r}{6,816}            & \multicolumn{1}{r}{485}            & 983                                & 2.36                                                                                    & 3.67                                                                                     \\
{\color[HTML]{CB0000} \textbf{SeqTopV2}}     & \multicolumn{1}{r}{94,458}           & \multicolumn{1}{r}{13,477}          & 6,095                               & 1.2                                                                                     & 1.98                                                                                     \\ \hdashline
\textbf{Total}                               & \multicolumn{1}{r}{\textbf{159,476}} & \multicolumn{1}{r}{\textbf{17,946}} & \textbf{11,618}                     & \multicolumn{1}{l}{}                                                                    & \multicolumn{1}{l}{}                                                                     \\ \bottomrule 
{\color[HTML]{009901} \textbf{ToolLLM-G1}}   & -                                   & -                                  & 197                                & 2.28                                                                                    & \multicolumn{1}{c}{-}                                                                    \\
{\color[HTML]{009901} \textbf{ToolLLM-G2} }   & -                                   & -                                  & 197                                & 2.55                                                                                    & \multicolumn{1}{c}{-}                                                                    \\
{\color[HTML]{009901} \textbf{ToolLLM-G3}}   & -                                   & -                                  & 97                                 & 2.91                                                                                    & \multicolumn{1}{c}{-}                                                                    \\
{\color[HTML]{009901} \textbf{API Bank-1}}   & -                                   & -                                  & 386                                & 1.65                                                                                    & 2.25                                                                                     \\
{\color[HTML]{009901} \textbf{API Bank-2}}   & -                                   & -                                  & 71                                 & 1.34                                                                                    & 2.44                                                                                     \\
{\color[HTML]{009901} \textbf{ToolBench-HS}} & -                                   & -                                  & 100                                & 7.01                                                                                    & 0.86                                                                                     \\
{\color[HTML]{009901} \textbf{ToolBench-B}}  & -                                   & -                                  & 120                                & 9.45                                                                                    & 0.89                                                                                     \\
{\color[HTML]{009901} \textbf{SeqToolQA}}    & -                                   & -                                  & 358                                & 2.42                                                                                    & 1.45                                                                                     \\
{\color[HTML]{009901} \textbf{ToolAlpaca}}   & -                                   & -                                  & 211                                & 1.38                                                                                    & 2.01                                                                                     \\ \hdashline
\textbf{Total}                               & -                                   & -                                  & \textbf{1737}                      &                                                                                         &                                                                                          \\ \bottomrule
\end{tabular} 
\caption{\sysname{} Datasets Statistics: datasets colored in {\color[HTML]{CB0000}red} are used for training and in-domain testing, while the {\color[HTML]{009901}green} ones are used for OOD testing only}
\label{tab:dataset_stat}
\end{table}

\begin{figure*}[hbt!]
\begin{center}
\includegraphics[width=0.95\textwidth]{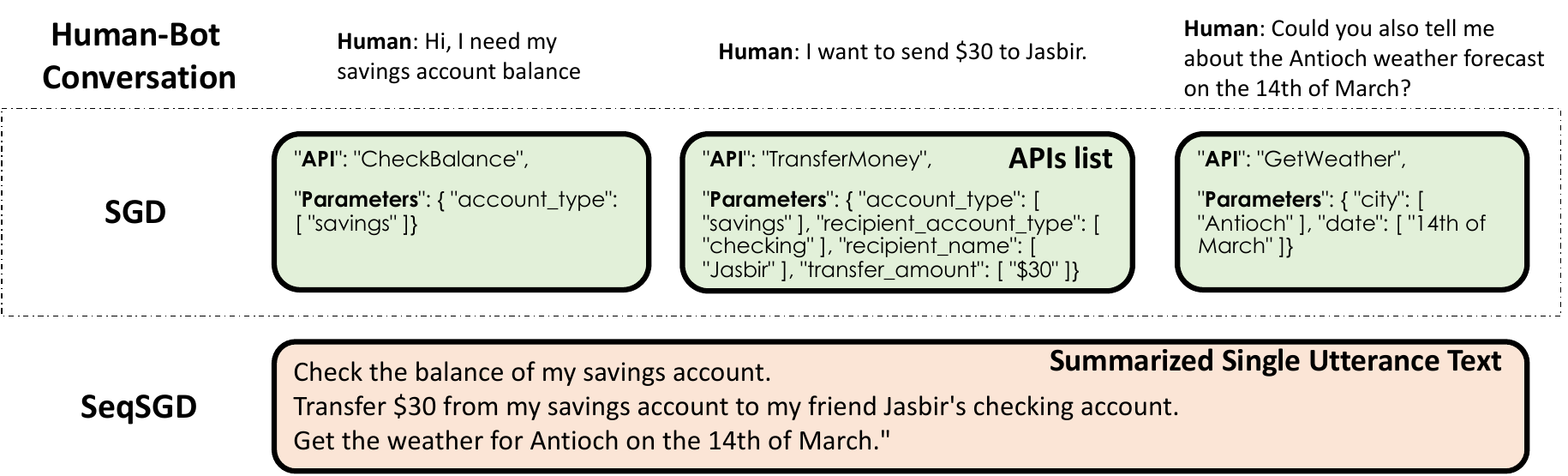}
\caption{Example of the creation process of seqSGD. Starting from the list of APIs, we use few-shot prompting to generate the summarized single utterance.}
\label{sgd_ex}
\end{center}
\end{figure*}

\section{\sysname{} Dataset Curation}

We focus on the setting where the input is a single natural language utterance and the output is a sequence of API calls with their parameter names and values. \sysname{} consists of datasets created via the following three approaches: (1) Language Model Assisted approach where prompts are used based on existing API outputs, (2) a grammar rule-based approach to convert existing semantic parsing and personal assistant notations into API data, and (3) off-the-shelf datasets. Table~\ref{tab:dataset_stat} depicts the statistics of each dataset and the details of the approach/dataset is below.


\subsection{Language Model Assisted Generation}

\textbf{SeqSGD}: We created SeqSGD, a dataset based on Schema-Guided Dialogue (SGD)  \cite{rastogi2020towards} dataset tailored towards API sequence evaluation. SGD contains about 20k annotated conversations between a human and a virtual assistant. These dialogues consist of engagements with various services and APIs across 20 domains, including banks, events, media, calendar, travel, and weather. To convert this dataset, for each conversation, we prompted a pretrained FLAN-T5-XXL\footnote{\url{https://huggingface.co/google/flan-t5-xxl}} model to convert each API into a request in natural language. We then append the corresponding text of each API to generate the summarized utterance. Figure \ref{sgd_ex} shows an example. We did not summarize the conversation itself, because it also contains API execution results, and using this as input to a summarization model resulted in many noisy details. To make sure the generated text captures all APIs and parameter values, we post-process the dataset to remove any examples where the utterance does not correctly reflect the ground truth APIs. As a result of this process, we generated 6.8K train, 1.1K validation, and 1.6K test examples having an average API count of 2.44 and an average parameters count per API of 3.5.     

\begin{figure*}[t]
\begin{center}
\includegraphics[width=0.95\textwidth]{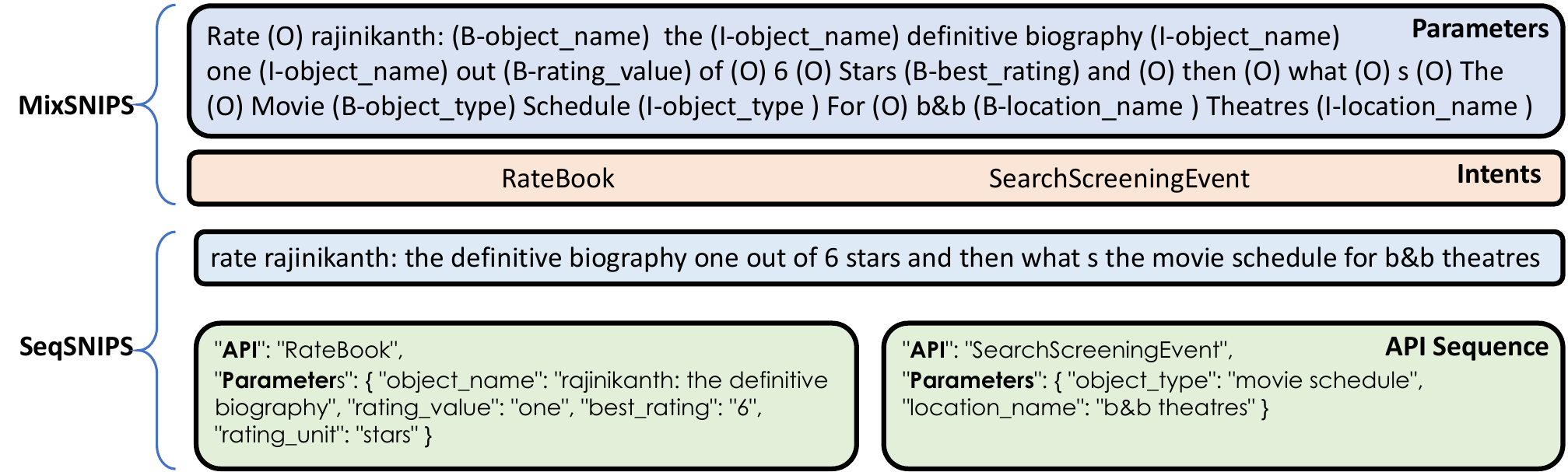}
\caption{Example of how SeqSNIPS is created. Using a natural language utterance from MixSNIPS and the flat list of slots, we convert it into a sequence of API calls, each with a dictionary of parameter names and values.}
\label{snips_ex}
\end{center}
\end{figure*}

\noindent
\textbf{SeqMultiWoz}: MultiWoz~\cite{ye2021multiwoz} is another multi-domain task-oriented dialogue dataset. Following the same process of curating SeqSGD from the SGD dataset, we created SeqMultiWoz, another API dataset based on MultiWoz. The resulting dataset includes about 6.8k train, 485 validation, and 1k test samples with an average API count of 2.36 and an average parameters count per API of 3.67.  

\subsection{Grammar-Based Generation}

\textbf{SeqATIS and SeqSNIPS}: ATIS \cite{hemphill-etal-1990-atis} is a collection of human-annotated queries about flights. Queries ask for information such as flight numbers, airports, dates, and other relevant details. 
The dataset also provides a range of semantic labels, including intent and slot values. Intents are the overall goals of the queries, such as ``flight query'' or ``airfare query'' while slot values are the specific pieces of information that are being requested, such as ``departure city'' or ``arrival time''. SNIPS \cite{coucke2018snips}  is another dataset focused on voice assistants. It consists of human-annotated queries that cover various domains such as weather, music, and calendar. 

MixATIS and MixSNIPS are multi-intent datasets ~\cite{qin2020agif} built based on ATIS and SNIPS, respectively. It was created by collecting sentences from the ATIS/SNIPS dataset and connecting them with conjunctions, such as ``and''. The resulting data had sentences with 1-3 intents at a probability of 30\%, 50\%, and 20\%, respectively. 

One issue with using these two datasets for API calling is that they do not indicate which parameters should be associated with which API. 
For example, as Figure \ref{snips_ex} shows, the original MixSNIPS dataset only evaluates model's ability to detect the two gold intents and which segments of the text are the target slots. 
To convert MixATIS and MixSNIPS datasets to a sequence of API calls, we divided utterances back to their original single intent utterances to get the corresponding parameters for each API. We then parsed its IOB (Inside/Outside/Beginning) parameter notations to generate the list of API parameter names and values. Now, we merge the utterances back along with the APIs and their parameters to get the sequence of API calls. In this way, we have curated SeqATIS and SeqSNIPS from MixATIS and MixSNIPS, respectively. In SeqATIS, we have around 11.5k train, 700 validation, and 800 test examples having an average API sequence length of 2.13 and an average parameter count per API of 4.85. Whereas, SeqSNIPS consists of around 40k train, 2.2k validation, and 2.2k test samples with an average API sequence length of 1.96 and an average parameters count per API of 5.06.

\noindent
\textbf{SeqToolQA}: ToolQA \cite{zhuang2023toolqa} is an open-source dataset designed for tool-augmented LLMs evaluation. The dataset tries to address the issue with current tool-based evaluation methods which do not distinguish between questions that can be answered using LLMs’ internal knowledge and those that require external information through tool use. 
To do so, the datasets come with 8 in-dependant datasets (e.g. Kaggle Flights, Coffee, Yelp, Airbnb, etc.) and a set of questions that can be answered only by querying those datasets, hence ruling out the internal knowledge of LLMs. The dataset is provided as a set of template-based questions with their final answers. 

However, ToolQA does not provide the intermediate set of steps (tools) that need to be executed to generate the answer to the given questions. The listing below shows an example:

\begin{lstlisting}[language=json]
{
 "qid": "easy-flight-0003", 
 "question": "What was the departure time of the 
       DL891 flight from SEA to LAX on 2022-01-22?",
 "answer": "11:54"
}
\end{lstlisting}
Too address this issue, we propose SeqToolQA where we used the provided implementation of ToolQA on how each template question is answered and transformed into a corresponding set of APIs. We abstracted 17 different APIs covering 6 domains and created a total of 358 examples for testing purposes. We show an example below:

\begin{lstlisting}[language=json,firstnumber=1]
{
 "qid": "easy-flight-0000",
 "question": "What was the departure time of the UA5480 flight from ORD to HSV on 2022-07-06?",
 "apis": [
    "LoadDB[DBName=flights]",
    "FilterDB[Origin=ORD,  Dest= HSV, FlightDate= 2022-07-06, Flight_Number_Marketing_Airline=5480, IATA_Code_Marketing_Airline=UA]",
    "GetValue[ValueName=DepTime]"
    ],
 "answer": "18:11"
}
\end{lstlisting}

\noindent
\textbf{SeqTopV2}: Topv2 \cite{chen-etal-2020-low-resource} is a multi-domain task-oriented semantic parsing dataset comprising examples from eight domains; alarm, event, messaging, music, navigation, reminder, timer, and weather. The total dataset consists of 180k samples, randomly divided into training, development, and test sets for each domain.
We followed a straightforward approach to convert this dataset into APIs using its intents ``IN:`` and slot ``SL:`` notations. Note that this dataset genuinely has a sequence of APIs that has to be followed. In the example ``Remind me to email Joy about the details with the new client tonight``, we transform it into two APIs; ``SEND\_MESSAGE`` and ``CREATE\_REMINDER`` where ``CREATE\_REMINDER`` has prerequisite for ``SEND\_MESSAGE``.   

The original dataset had a lot of "UNSUPPORTED" notations for examples where there is no matching intent. We excluded these cases from our API dataset. Along with them, we also removed the samples that had duplicate utterances, ambiguous intents, and analogous slot names inside the same intent. We call the resulting dataset SeqTopV2, and it has 94K, 13.5K, and 6K for training, development, and testing splits, respectively. 

\subsection{Off-the-Shelf Usage}

\textbf{ToolBench}: This is a subset of ToolBench \cite{xu2023tool} focused on two domains, HomeSearch and Booking. We did not do any transformation to these datasets and rather used it ``as-is'', since they are already in API form. 



\noindent
\textbf{ToolLLM}~\cite{qin2023toolllm} proposes an instruction-tuning tools dataset and a tool-augmented LLM model based on LLaMA-2-7B. The dataset is created synthetically based on ChatGPT and a collection of 16K APIs from RapidAPI.  The dataset include three subsets; G1,G2, G3 which refer to single-tool, intra-category multi-tool and intra-collection multi-tool data, respectively. We used those three subsets above as-is for out-of-distribution model evaluation. 

\noindent
\textbf{API Bank}~\cite{li2023apibank}: is a benchmark  designed for evaluating tool-augmented LLMs. It includes 314 tool-use dialogues with 753 API calls to assess the existing LLMs' capabilities in planning, retrieving, and calling APIs. It also comes with a larger training dataset of 1.8K dialogues and a model trained on it (Lynx) initialized from Alpaca. In this paper, we use the the test sets of API Bank for out-of-distribution evaluation. Since this is a dialoge with tools being called in between, we divided each dialogue into multiple test examples where each example include a conversation and a sequence of APIs needed thus far. 

\noindent
\textbf{ToolAlpaca} \cite{toolalpaca}: ToolAlpaca is a training and evaluation benchmark that is automatically curated through a multi-agent simulation environment using ChatGPT and GPT-3.5. The corpus contains 3,938 tool-use instances from more than 400 real-world tool APIs spanning 50 distinct categories. Although it comes with a training set, in this paper, we have only used the test set for out-of-distribution evaluation.

%% file: sections/5-experiments.tex
\section{Experiments and Results}

\begin{table*}
\scriptsize
\centering
\resizebox{2.1\columnwidth}{!}{%
\begin{tabular}{c|c|ccccc|c}
\toprule
FT Types                       & Models                  & {\color[HTML]{CB0000} \textbf{SeqATIS}} & {\color[HTML]{CB0000} \textbf{SeqSNIPS}} & {\color[HTML]{CB0000} \textbf{SeqSGD}} & {\color[HTML]{CB0000} \textbf{SeqMultiWOZ}} & {\color[HTML]{CB0000} \textbf{SeqTopV2}} & \textbf{Weighted Avg.}             \\ \midrule
                               & \textbf{Falcon-180B}    & 0.15 | 0.02 | 0.15                      & 0.39 | 0.07 | 0.40                        & 0.21 | 0.06 | 0.21                     & 0.44 | 0.25 | 0.45                          & 0.08 | 0.00 | 0.09                          & 0.19 | 0.04 | 0.20           \\
                               & \textbf{LLaMA-2-70B}    & 0.10 | 0.01 | 0.11                       & 0.26 | 0.03 | 0.27                       & 0.10 | 0.02 | 0.10                       & 0.23 | 0.12 | 0.25                          & 0.04 | 0.00 | 0.04                          & 0.11 | 0.02 | 0.12          \\
\multirow{-3}{*}{No FT}        & \textbf{ToolLLaMA-2-7B} & 0.29 | 0.03 | 0.32                      & 0.49 | 0.04 | 0.47                       & 0.43 | 0.05 | 0.45                     & 0.78 | 0.40 | 0.79                           & 0.07 | 0.00 | 0.08                          & 0.27 | 0.05 | 0.28          \\ \midrule
                               & \textbf{FLAN-T5-XXL}    & 0.92 | 0.72 | 0.92                      & 0.97 | 0.90 | 0.97                        & 0.98 | 0.69 | 0.98                     & 1.00 | 0.99 | 1.00                                & 0.96 | 0.83 | 0.96                       & 0.97 | 0.83 | 0.97          \\
                               & \textbf{StarCoder-15B}      & 0.99 | 0.84 | 0.99                      & 0.96 | 0.87 | 0.96                       & 0.98 | 0.67 | 0.98                     & 1.00 | 0.99 | 1.00                                & 0.95 | 0.78 | 0.95                       & 0.96 | 0.80 | 0.96           \\
                               & \textbf{Facon-40B}      & 0.92 | 0.70 | 0.92                       & 0.97 | 0.89 | 0.97                       & 0.96 | 0.62 | 0.96                     & 1.00 | 0.97 | 1.00                                & 0.90 | 0.56 | 0.91                        & 0.93 | 0.67 | 0.94          \\
\multirow{-4}{*}{\shortstack{FT w. dataset \\ mentioned \\ in the column}} & \textbf{MPT-30b}        & 0.96 | 0.81 | 0.96                      & 0.97 | 0.90 | 0.97                        & 0.98 | 0.68 | 0.98                     & 1.00 | 0.98 | 1.00                                & 0.96 | 0.84 | 0.97                       & \textbf{0.97 | 0.84 | 0.97} \\ \midrule
                               & \textbf{FLAN-T5-XXL}    & 0.94 | 0.72 | 0.94                      & 0.96 | 0.89 | 0.97                       & 0.98 | 0.70 | 0.98                      & 1.00 | 0.97 | 1.00                                & 0.97 | 0.87 | 0.97                       & 0.97 | 0.85 | 0.97          \\
                               & \textbf{StarCoder-15B}      & 0.98 | 0.81 | 0.98                      & 0.96 | 0.86 | 0.96                       & 0.98 | 0.67 | 0.98                     & 1.00 | 0.96 | 1.00                                & 0.96 | 0.83 | 0.96                       & 0.97 | 0.82 | 0.97          \\
                               & \textbf{LLaMA-2-7B}      & 0.92 | 0.70 | 0.92                       & 0.96 | 0.88 | 0.97                       & 0.97 | 0.65 | 0.97                     & 1.00 | 0.97 | 1.00                                & 0.96 | 0.85 | 0.97                       & 0.96 | 0.83 | 0.97          \\
                               & \textbf{Facon-40B}      & 0.90 | 0.67 | 0.90                        & 0.96 | 0.87 | 0.97                       & 0.94 | 0.62 | 0.94                     & 1.00 | 0.94 | 1.00                                & 0.93 | 0.66 | 0.93                       & 0.94 | 0.72 | 0.94          \\
\multirow{-5}{*}{FT w. all data} & \textbf{MPT-30B}        & 0.94 | 0.77 | 0.94                      & 0.97 | 0.90 | 0.97                        & 0.98 | 0.70 | 0.98                      & 1.00 | 0.97 | 1.00                                & 0.97 | 0.87 | 0.97                       & \textbf{0.97 | 0.85 | 0.97} \\ \bottomrule
\end{tabular}
}
\caption{Evaluation Results on \textbf{In-Distribution} datasets. Scores are shown in the following format: \textbf{API-F1 | Parameter-F1 | LCS-F1}. The weighted average scores are calculated using the number of test samples in Table \ref{tab:dataset_stat}.}
\label{tab:in_domain_eval}
\end{table*}

\subsection{Baselines}

In our experiments, we have used 9 open sourced models as baselines: (1) LLaMA-2-70B \cite{touvron2023llama}, (2) Falcon-180B \cite{almazrouei2023falcon}, (3) LLAMA-2-7B \cite{touvron2023llama}, (4) FLAN-T5-XXL \cite{flan_t5}, (5) Falcon-40B \cite{almazrouei2023falcon}, (6) StarCoder-15B \cite{li2023starcoder}, (7) MPT-30B \cite{mpt7b}, (8) ToolLLaMA-2-7B \cite{qin2023toolllm}, and (9) Lynx-7B \cite{li2023apibank}.
We tested these models in three settings; (1) few shot testing: we evaluated LLAMA-2-70B, Falcon-180B, and ToolLLaMA-2-7B in a 3 shot mode; (2) Instruction fine-tuning on target dataset: we consider this setting for FLAN-T5-XXL, StarCoder-15B, Falcon-40B, and MPT-30B; and (3) Instruction fine-tuning on combined datasets: we evaluated this setting for all models in (2) along with LLaMA-2-7B to evaluate whether we can get a single model trained on the plurality of all datasets and still perform well on each individual test set. For the OOD experiments, we have used all the fine-tuned models from (3) in conjunction with the ToolLLaMA-2-7B and Lynx-7B which are already fine-tuned with the ToolLLM and APIBench data, respectively.

\subsection{Instruction Tuning}
In all experiments, we have used the same instructions for training and testing. We show below the instruction template. Only when evaluating non-fine-tuned models or for the OOD experiments, we also provide 3 ICL examples via the part ``Here are some examples: {ICL\_EXAMPLES}'' in the instruction) and remove it otherwise.

\begin{lstlisting}[language=json,firstnumber=1]
### Instruction Template with ICL Examples ###
Given the APIs and Slots below, sequence them in the order in which they have to be called to answer the following query. 
Possible APIs: {INTENT_LIST} 
Possible Slots: {SLOT_LIST}
Here are some examples: {ICL_EXAMPLES}
Query: {QUERY}
Answer: 
\end{lstlisting}

\subsection{Settings and Parameters:}
We used QLoRA \cite{dettmers2023qlora} to fine-tune all our models. While fine-tuning the models on targeted datasets, we made sure that the model saw 100k samples in the training process. In combined data training, we fine-tuned the models for $2$ epochs over the cumulated datasets. In both cases, the batch size was $1$ with gradient accumulation steps of $8$ and a learning rate of $5e^{-5}$.

\subsection{Metrics}
To perform a fine-grained evaluation of the generated responses, we use two kinds of evaluation - standard information retrieval metrics (precision, recall, and f1 scores) and Longest Common Subsequence (LCS).  We report F1 APIs and F1 slots/Parameters to compute the F1 scores by comparing the predicted APIs with the gold ones and the predicted parameters of each API with its gold counterparts. To also measure the model's ability to follow the sequence of API calls as dictated by the given natural language query, we also report LCS by comparing the gold sequence of API names and the sequence predicted by the model.  

\begin{table*}[hbt!]
\scriptsize
\centering
\resizebox{2.1\columnwidth}{!}{%
\begin{tabular}{c|ccccc|cc}
\toprule
                                             & \multicolumn{5}{c|}{\textbf{Fine-Tuned with all \sysname{} data}}                                                                           & \multicolumn{2}{c}{\textbf{Tool-Augmented LLMs}} \\ 
\multirow{-2}{*}{\textbf{Datasets}}          & \textbf{Falcon-40B} & \textbf{FLAN-T5-XXL}  & \textbf{MPT-30B}      & \textbf{LlaMA-2-7B} & \textbf{StarCoder-15B} & \textbf{ToolLLaMA-2-7B}    & \textbf{Lynx-7B}        \\ \midrule
{\color[HTML]{009901} \textbf{ToolLLM-G1}}   & 0.48 | 0.47              & -                          & 0.11 | 0.11                 & 0.07 | 0.07              & 0.32 | 0.32              & 0.12 | 0.12               & 0.43 | 0.44          \\
{\color[HTML]{009901} \textbf{ToolLLM-G2}}   & 0.48 | 0.47              & -                          & 0.24 | 0.23                 & 0.09 | 0.08              & 0.53 | 0.53              & 0.01 | 0.01               & 0.33 | 0.34          \\
{\color[HTML]{009901} \textbf{ToolLLM-G3}}   & 0.50 | 0.49               & -                          & 0.51 | 0.49                 & 0.19 | 0.20               & 0.49 | 0.48              & 0.16 | 0.16               & 0.50 | 0.50            \\
{\color[HTML]{009901} \textbf{API Bank-1}}   & 0.42 | 0.15 | 0.44       & 0.57 | 0.12 | 0.59         & 0.59 | 0.21 | 0.62          & 0.52 | 0.16 | 0.55       & 0.49 | 0.15 | 0.55       & 0.11 | 0.04 | 0.12        & 0.31 | 0.11 | 0.33   \\
{\color[HTML]{009901} \textbf{API Bank-2}}   & 0.37 | 0.16 | 0.39       & 0.51 | 0.11 | 0.53         & 0.49 | 0.20 | 0.52           & 0.38 | 0.14 | 0.40        & 0.45 | 0.13 | 0.48       & 0.05 | 0.03 | 0.05        & 0.19 | 0.09 | 0.20    \\
{\color[HTML]{009901} \textbf{ToolBench-HS}} & 0.95 | 0.77 | 0.92       & 0.42 | 0.16 | 0.43         & 0.91 | 0.76 | 0.80           & 0.69 | 0.41 | 0.54       & 0.90 | 0.81 | 0.89        & 0.59 | 0.35 | 0.56        & 0.77 | 0.55 | 0.76   \\
{\color[HTML]{009901} \textbf{ToolBench-B}}  & 0.89 | 0.76 | 0.76       & 0.36 | 0.09 | 0.33         & 0.85 | 0.72 | 0.78          & 0.76 | 0.49 | 0.56       & 0.44 | 0.34 | 0.40        & 0.73 | 0.52 | 0.63        & 0.57 | 0.47 | 0.48   \\
{\color[HTML]{009901} \textbf{SeqToolQA}}    & 0.27 | 0.02 | 0.28       & 0.18 | 0.00 | 0.19            & 0.48 | 0.02 | 0.51          & 0.50 | 0.00 | 0.53           & 0.24 | 0.01 | 0.26       & 0.14 | 0.00 | 0.14           & 0.27 | 0.02 | 0.29   \\
{\color[HTML]{009901} \textbf{ToolAlpaca}}   & 0.41 | 0.11 | 0.41 & 0.54 | 0.13 | 0.55 & 0.51 | 0.12 | 0.52 & 0.46 | 0.13 | 0.46 & 0.47 | 0.13 | 0.49 & 0.18 | 0.02 | 0.20 & 0.32 | 0.04 | 0.34      \\ \hdashline
\textbf{Weighted Avg.}                             & 0.47 | 0.21 | 0.46 & 0.42 | 0.09 | 0.43 & \textbf{0.49 | 0.23 | 0.49} & 0.41 | 0.16 | 0.40 & 0.44 | 0.17 | 0.46 & 0.18 | 0.09 | 0.18 & 0.37 | 0.14 | 0.38  \\
\bottomrule
\end{tabular}
}
\caption{Evaluation Results on \textbf{Out-of-Distribution (OOD)} dataset. Each scores are shown in the following format: \textbf{API-F1 | Parameter-F1 | LCS-F1}, except the ToolLLM datasets, which are API-only, so they do not have the Parameter-F1 score. All models are prompted with 3-shot examples. The weighted average scores are calculated using the number of test samples in Table ~\ref{tab:dataset_stat}.}
\label{tab:out_of_domain_eval}
\end{table*}

\subsection{In-Distribution Evaluation Results}

\textbf{No Fine-tuning evaluation}: 
The first experiment we did was to check how the state-of-the-art open LLMs perform in such a setting. In particular, we evaluated LLaMA-2-70B and Falcon-180B using 3-shot prompting. We also considered ToolLLaMA-2-7B \cite{qin2023toolllm}; a LLAMA-2-7B based model trained on API datasets generated using ChatGPT based on specifications from RapidAPIs. 
Table \ref{tab:in_domain_eval} shows the evaluation results on five in-distribution datasets: SeqATIS, SeqSNIPS, SeqSGD, SeqMultiWoz, and SeqTopV2. On all datasets, all three non-fine-tuned models seem to get some of the APIs correctly but fail to get the parameters to call such APIs. 

\noindent
\textbf{Fine-tuning on One Dataset}:
In this experiment, we fine-tune the baselines discussed above on each dataset and test it on the corresponding test split. We have evaluated four models here: FLAN-T5-XXL, StarCoder-15B, Falcon-40B, and MPT-30B. Ideally, this setting should give the best performance since the training data is focused towards one dataset and its APIs. As shown in Table \ref{tab:in_domain_eval}, all models achieved very good performance ($>$ 90\%) detecting the right APIs with the performance of all models reaching 100\% API-F1 scores for SeqMultiWoz dataset. We hypothesize that this high performance is because the number of APIs in most datasets is not very large (e.g., 12 APIs for SeqMultiWoz). Detecting the correct set of parameter names and values is a more challenging problem for most models with performance being the lowest for the SeqSGD dataset. The weighted average number suggests that the MPT-30B model is doing slightly better than the other models. 

\noindent
\textbf{Fine-tuning on All Training Datasets}: 
In this setting, we combine all training datasets and fine-tune the five baseline models on them. 
The goal of this experiment is to check if we can get a generic model that works well when tested on each individual dataset. 
We see in Table~\ref{tab:in_domain_eval}, on SeqATIS and SeqMultiWoz, models trained on the combined training data achieve lower performance compared to models trained on the individual dataset. Performance on SeqSNIPS was similar for both models, while models trained on the combined data achieved better performance on SeqSGD and SeqTopV2. The average scores suggest that all models achieved better performance when trained on the combined datasets compared with the single dataset training. 

\subsection{Out-Of-Distribution Evaluation}
To check the generality of the different API models, we measure the performance of all models on five out-of-distribution test sets; ToolLLM, API Bank, ToolBench, our SeqToolQA, and ToolAlpaca. Some test sets have subcategories, such as ToolLLM has G1, G2, and G3; API-Bank has L1 and L2; and ToolBench has HS and B for Home Search and Booking, respectively. In our test-suite, ToolLLM is the API-only test-set that does not have any parameters.   
In this experiment, we use the 5 models (i.e., FLAN-T5-XXL, StarCoder-15B, Falcon-40B, MPT-30B, and LLaMA2-7B) that are fine-tuned with our combined data. In addition to these models, we have also evaluated ToolLLaMA-2-7B from ToolLLM ~\cite{qin2023toolllm} and Lynx-7B from API-Bank ~\cite{li2023apibank}. In this experiment, we compare the performance with 3-shot in-context learning examples for all the models.

Table \ref{tab:out_of_domain_eval} showcases our OOD evaluation results. Our models that are fine-tuned with \sysname{} data perform better than other Tool/API augmented models. This is because our models achieve better generalizability as they have seen diverse sets of API data from different domains in the fine-tuning process. Falcon-40B and StarCoder-15B are showing better performance on our API-only test-set ToolLLM (we did not evaluate  FLAN-T5-XXL on ToolLLM due to max sequence limit), whereas FLAN-T5-XXL and MPT-30B are doing well on API-Bank and ToolAlpaca. Even though ToolLLaMA-2-7B and Lynx-7B are from ToolLLM and API-Bank respectively, still they are performing poorly. In their papers, they used different metrics, e.g., pass rate to determine how many times the model reached an answer, and win rate which uses ChatGPT as a judge. 
In both the ToolBench datasets, Falcon-40B is outperforming the others. On SeqToolQA, even though the models have scored some points in API detection, however, all the models performed poorly on detecting the parameter names and values, which leads to a low Parameter-F1 score. This is because the parameter values in SeqToolQA contain codes, SQL, math functions, etc., which models have not seen in any training data, and these special slot values are not trivial to generate by seeing only a few ICL examples.  

\subsection{Qualitative Studies}
We also performed extensive studies on the outputs generated by the models in our experiments. In this section, we are going to discuss our findings on the failed cases along with some common mistakes demonstrated by the models. We found, in most of the cases, that parameters names and values are the obvious reason for not doing well on slot detection in both in and out-of-distribution test sets. We provide samples for each case for better understanding. We would like to mention that most of the provided examples contain ``gold\_output'' and ``predicted\_output'' and we have shown only the error part (sub-string) from the original outputs for brevity.






\subsubsection{Unnormalized Slot-Values}
In an ideal scenario, the parameter values should be extracted exactly from the query by the models while generating the texts. However, sometimes, the model extracts the sub-part of it or represents it in a different form after extracting it. In a human evaluation, we would consider the generated text matches the gold, although while doing it programmatically it's showing a mismatch and we have not found a good way to normalize the parameter values. The following examples capture some of the unnormalized parameter value mismatches. In the first example, the month and the day in the predicted output are repeated. The predicted output on the second one contains only the city name, whereas the gold contains the city and the state. In the final example, even if the intent and slot values are correct, they have used different parameter formats to represent it.  We plan to investigate further these issues, but we keep it for future work. 

\begin{lstlisting}[language=json,firstnumber=1]
### SeqSGD ###
{  "gold": "March 3rd, this Sunday",
   "predicted": "March 3rd, the 3rd"
}
{  "gold": "NYC, New York",
   "predicted": "NYC"
}
### ToolBench ###
{  "gold": "Date(3, 9, 2023)",
   "predicted": "Date(year=2023, month=3, day=9)"
}
\end{lstlisting}


\subsubsection{Semantically similar slot-names in API Specification}
In our instructions, we provide the possible list of APIs and parameters to be used by the model while answering the queries. We extract these APIs and parameters from the original dataset's API specifications, if any. However, we found in some cases the parameter names are semantically very similar across the datasets. 
Here are some examples from the SeqSGD dataset: (1) \textit{leaving\_date} and \textit{departure\_date}; (2) \textit{leaving\_time} and \textit{departure\_time}; (3) \textit{origin}, \textit{from\_city}, and \textit{from\_location}; and (4)
\textit{destination}, \textit{to\_city}, and \textit{to\_location}. Now, it often happens that the parameter values are correct in the generated text but the parameter names do not exactly match with the gold, even if they are very close. Following are some examples of such cases.

\begin{lstlisting}[language=json,firstnumber=1]
### SeqSGD ###
{ "gold": "destination_airport = ATL",
  "predicted": "destination = ATL"
}
{ "gold": "show_type = imax",
  "predicted": "theater_name = imax"
}
### SeqATIS ###
{  "gold": "cuisine = souvlaki",
   "predicted": "served_dish = souvlaki"
}
\end{lstlisting}

%% file: sections/limitations.tex
\section{Limitations and Risks}

A limitation of our benchmark, \sysname{}, is that it does not deal with environment interactions for an API agent. In future work, it will be interesting to explore this setting of an embodied agent, where the API calls effect changes in the grounded environment. Further, our benchmark is focused on English API commands, and in the future, it will be interesting to develop a multilingual \sysname{}. We do not perceive any risks associated with our work.